\documentclass[conference]{IEEEtran}
\IEEEoverridecommandlockouts
\usepackage{cite}
\usepackage{amsmath,amssymb,amsfonts}
\usepackage{algorithmic}
\usepackage{graphicx}
\usepackage{textcomp}
\usepackage{xcolor}
\usepackage{multirow}
\usepackage{colortbl}
\usepackage{amsmath}
\usepackage{amssymb}
\usepackage{cite}
\usepackage[ruled,vlined]{algorithm2e}

\newcommand{\comment}[1]{}
\def\BibTeX{{\rm B\kern-.05em{\sc i\kern-.025em b}\kern-.08em
    T\kern-.1667em\lower.7ex\hbox{E}\kern-.125emX}}

\makeatletter

\begin{document}

\title{Detection of Morphed Face Images Using Discriminative Wavelet Sub-bands}

\author{\IEEEauthorblockN{Poorya Aghdaie, Baaria Chaudhary, Sobhan Soleymani, Jeremy Dawson, Nasser M. Nasrabadi}
\IEEEauthorblockA{\textit{West Virginia University} \\
}}

\maketitle

\begin{abstract}
This work investigates the well-known problem of morphing attacks, which has drawn considerable attention in the biometrics community. Morphed images have exposed face recognition systems' susceptibility to false acceptance, resulting in dire consequences, especially for national security applications. To detect morphing attacks, we propose a method which is based on a discriminative 2D Discrete Wavelet Transform (2D-DWT). A discriminative wavelet sub-band can highlight inconsistencies between a real and a morphed image. We observe that there is a salient discrepancy between the entropy of a given sub-band in a bona fide image, and the same sub-band's entropy in a morphed sample. Considering this dissimilarity between these two entropy values, we find the  Kullback-Leibler divergence between the two distributions, namely the entropy of the bona fide and the corresponding morphed images. The most discriminative wavelet sub-bands are those with the highest corresponding KL-divergence values. Accordingly, 22 sub-bands are selected as the most discriminative ones in terms of morph detection. We show that a Deep Neural Network (DNN) trained on the 22 discriminative sub-bands can detect morphed samples precisely. Most importantly, the effectiveness of our algorithm is validated through experiments on three datasets: VISAPP17, LMA, and MorGAN. We also performed an ablation study on the sub-band selection.
\end{abstract}


\section{Introduction}
Morphing attack detection is of great significance in high-throughput border control applications. According to the CIA triad model, consisting of three main components, confidentiality, integrity, and availability of secure systems, morphed images violate the integrity of verification systems. A morphed image is generated using genuine face images from two different individuals. Because the resulting morphed image inherits characteristics of both subjects, it can be verified against both real subjects. Morphed images are generated using two approaches. In the first approach \cite{makrushin2017automatic, seibold2018accurate, seibold2017detection}, two real face images are alpha blended in order to create a morphed image. To eliminate the ghosting effects in the morphed image, the average of the landmarks in both real images is used as the resulting landmark of the morphed image.  In the second approach introduced in \cite{damer2018morgan}, a generative model, that is a Generative Adversarial Network (GAN), is trained to synthesize morphed images. Morph detection algorithms can be grouped into two main categories: single and differential morph detection. In the first category, an image under investigation is labeled as morphed or bona fide image, which is known as single image morph detection. In differential morph detection, a subject's image is compared with a live capture of the subject, and information from both images is used to detect morphed counterfeits.

\begin{figure}[t]
\begin{center}
\includegraphics[width=.95\linewidth]{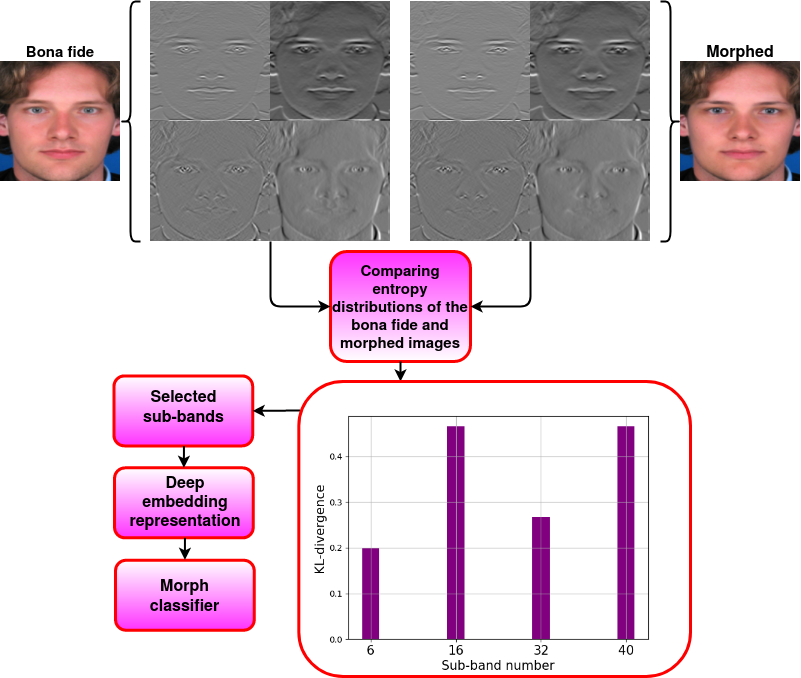} 
\end{center}
  \caption{A bona fide and a morphed image along with the four corresponding wavelet sub-bands. Using all the bona fide and morphed images in the dataset, 48 pairs of entropy distributions are found for bona fide and morphed images. Given a sub-band, dissimilarity between the two entropy distributions represents how discriminative that sub-band is with respect to morph detection. In the figure, sub-bands 16 and 40 are more discriminative than 6 and 32. A deep classifier is trained using the selected informative sub-bands.}
\end{figure}

To detect morphed images, some of the previous research efforts employ hand-crafted features such as  Binarized Statistical Image Features (BSIF) \cite{kannala2012bsif}, Scale Invariant Feature Transform (SIFT) \cite{lowe1999object}, Speeded Up Robust Features (SURF) \cite{bay2006surf}, (Local Binary Patterns Histogram) LBPH \cite{ojala1994performance}, Fused Local Binary Pattern (FLBP), and Histogram of Gradianets (HOG). Recently, Deep Neural Networks (DNNs) have proved to be promising in detecting morphed images \cite{debiasi2019detection, venkatesh2020detecting}. Thus far, no wavelet-based morph detection algorithm has been proposed. In this work, we propose a single image morph detector which can distinguish between a bona fide and a morphed face image. To do so, we train a deep neural network with a small number of selected discriminative wavelet sub-bands that are chosen according to the following criterion: the relative entropy between the entropy distribution of real faces and morphed faces is found for each of the wavelet sub-bands. The higher the value of the relative entropy for a given sub-band, the more discriminative that sub-band is for the task of classification. Fig. 1 depicts our morph detection mechanism. Please note that only four wavelet sub-bands of a bona fide and its corresponding morphed image are selected in the figure for representation purpose. However, we consider all images in a given dataset to find the histogram of entropy for each sub-band. Experiments on three datasets, i.e., VISAPP17 \cite{makrushin2017automatic}, MorGAN\cite{damer2018morgan}, and LMA \cite{damer2018morgan} verifies the performance of our morph detector. Standard quantitative measures, set forth by ISO/IEC 30107-3 \cite{iso30107}, are used to evaluate the effectiveness of our proposed method. The first measure is Attack Presentation Classification Error Rate (APCER), which is the percentage of morphed images that are classified as bona fide. The second measure is Bona Fide Presentation Classification Error Rate (BPCER), which represents the percentage of bona fide samples that are classified as morphed. If we label the morphed class as positive and the bona fide class as negative , APCER, and BPCER are equivalent to false negative rate and false positive rate, respectively. 
The contributions of this paper are as follows: the most discriminative wavelet sub-bands are selected based on the KL-divergence between the two entropy distributions of both real and morphed images for the wavelet sub-bands. A DNN is trained using the selected informative wavelet sub-bands to detect morphed images. Finally, an ablation study is performed to show the effectiveness of our sub-band selection scheme for tackling detecting morph attacks.
\begin{figure*}[t]
\begin{center}
\includegraphics[width=.95\linewidth]{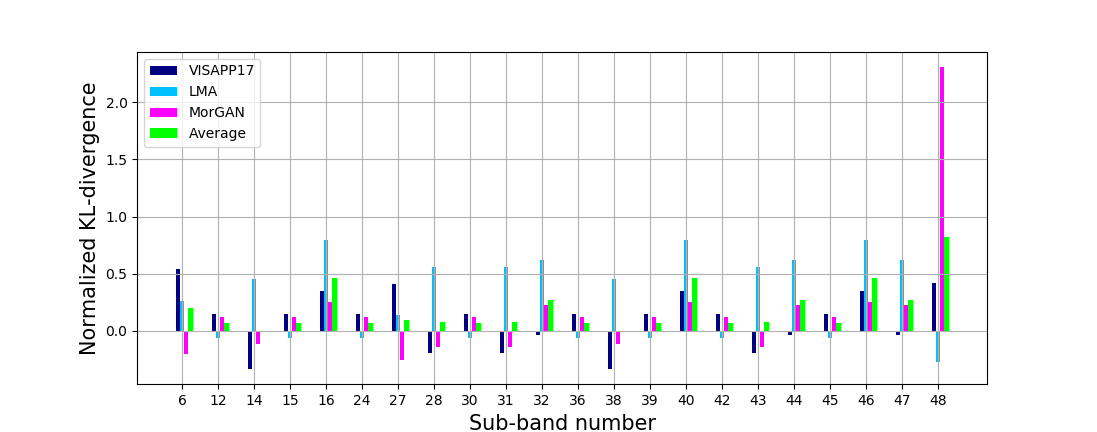} 
\end{center}
  \caption{Zero-meaned KL-divergence values in the top 22 most discriminative wavelet sub-bands for three datasets: VISAPP17, LMA, and MorGAN. The zero-meaned average of the KL-divergence values in each sub-band, as related to the three datasets, is represented in green.}
\end{figure*}

\section{Related Work}

 Some morph detection techniques use hand-crafted features for training a classifier to identify morphed samples \cite{scherhag2018towards, scherhag2018morph, damer2018morgan}. In addition, deep embedding features extracted using off-the-shelf DNNs can be utilized for training a morph detector\cite{raghavendra2017transferable, damer2018morgan, scherhag2018morph}. Damer et al. \cite{damer2018morgan} employ GANs for generating morphed face images. An SVM is trained for the task of morph detection using the crafted LBPH, as well as the extracted deep embedding feature vectors. One of the research efforts \cite{scherhag2019detection} adopts Photo Response Non-Uniformity (PRNU) to distinguish between real and morphed images. FLBP in color channels of HSV and YCbCr color spaces are studied in \cite{raghavendra2017face} as a method for detecting morphed images. One of the most important aspects of the morphing attacks is carefully selecting two bona fide subjects's face images such that the morphing attack looks highly photo realistic. In \cite{damer2019detect}, the morph detection is investigated when the morphed images are generated using three different pairing protocols: (1) two similar images for morphing, (2) two random images, and (3) two dissimilar images. 
As a holistic approach for morph detection, fusion of the above-mentioned algorithms can be considered. In \cite{scherhag2018morph}, two SVMs are trained using two different textures descriptors: LBPH, and BSIF. Another SVM is trained with the HOG, and deep embedding features are used to train another SVM. To integrate all approaches, the  resulting scores from all the detectors are fused. Another work \cite{venkatesh2020detecting} employs a denoised version of an image to find the residual noise of the image which can be utilized for identifying morphed samples. The paper aggregates several denoised versions of an image in the wavelet domain. Disentanglement of appearance and landmark is another method proposed for differential morph detection \cite{soleymani2021mutual}. Interestingly, reflection inconsistencies are also employed to detect morphing attacks \cite{seibold2018reflection}.

\section{Our Framework}

We employ undecimated 2D wavelet decomposition to address morphing attacks. Shannon entropy and Kullback-Liebler divergence \cite{kullback1951information} are utilized to identify the optimal discriminative sub-bands. In particular, the Shannon entropy \cite{shannon1948mathematical} is used to measure embedded information in each sub-band of the wavelet decomposition. Since most of the morphing pipeline artifacts lie in the high frequency spectrum, we do not consider the Low-Low (LL) sub-band of the first level of decomposition to be decomposed further. Instead, the Low-High (LH), High-Low (HL), and High-High (HH) sub-bands are decomposed . After 3-level uniform decomposition, 48 sub-bands are obtained, for all of which the Shannon entropy is computed, and the distribution of the entropy is obtained for both real and morphed images for the three training datasets. The Kullback-Leibler divergence (relative entropy) is calculated between the entropy distribution of real and morphed sub-bands for each of the 48 sub-bands, and these 48 relative entropy values are sorted from highest to lowest. A final subset composed of 22 optimal discriminative sub-bands are selected that are used to train a DNN to detect morphed samples. As for the DNN, we employ a pre-trained Inception Resnet v1 architecture as our binary classifier.

\subsection{Sub-band Selection Based on KL Divergence of Entropy Distributions}
The pivotal point here is to distinguish morphed samples by leveraging the most discriminative sub-bands. To do so, we find the histograms of entropy of all 48 sub-bands for both bona fide and morphed images in the three datasets. Accordingly, 96 distributions are estimated using the histograms from the 48 sub-bands of both the bona fide and morphed presentations. The term ${\hat{f}}_{b_i}$ represents the estimated distribution for the $i^{th}$ sub-band pertinent to the bona fide images, and similarly, ${\hat{f}}_{m_i}$ represents the estimated distribution for the $i^{th}$ sub-band pertinent to the morphed images. The dissimilarity of the two probability distribution functions, namely $({\hat{f}}_{b_i},{\hat{f}}_{m_i})$ are calculated for all 48 sub-bands. The KL-divergence is the metric we employ to assess the dissimilarity between the distributions. 

In order to select the most discriminative sub-bands, the KL-divergence values of each dataset are first normalized by removing the mean. The values are normalized to enable comparison of the distributions across the three datasets. Then, the zero-meaned values are averaged over the three datasets for each sub-band. The higher the KL-divergence value for a single sub-band, the more informative and discriminative that sub-band is in terms of classification. By choosing the sub-bands that are based on the highest average KL-divergence values from all datasets instead of each dataset separately, we can find the sub-bands that are discriminative across the datasets, not just for a specific morphing technique. Fig. 2 shows the distribution of the zero-meaned KL-divergence values related to the 22 most discriminative wavelet sub-bands for the three morphed datasets, and their average values. Algorithm 1 illustrates our sub-band selection mechanism, in which $H(.)$ represents the entropy function.

\begin{algorithm}[t]

\small
\SetAlgoLined
    \SetKwInOut{Input}{Input}
    \SetKwInOut{Output}{Output}
    \Input{Bona fide and Morphed Images}
    \Output{A Set of Indices for Informative Sub-bands  }
    $\mathbb{I}=\{\}$   \tcp*[l]{index of sub-bands} 
    \For{$i=1$ to 48\tcp*[l]{sub-bands} }{
        \For{$j=1$ to 3\tcp*[l]{datasets} }
        {
        ${\hat{f}}_{b_{ij}} \leftarrow distribution(H(S_{b_{ij}}))$
        ${\hat{f}}_{m_{ij}} \leftarrow distribution(H(S_{m_{ij}}))$
        $\mathbb{K}_{ij} \leftarrow D_{KL}({\hat{f}}_{b_{ij}}\|{\hat{f}}_{m_{ij}})$
        }
    }
    \For{$i=1$ to 48}{
    ${\Bar{\mathbb{K}}}_i \leftarrow \underset{j}{avg}(\mathbb{K}_{ij})$\\
    \If{$\Bar{\mathbb{K}}_i > threshold$}{
    $\mathbb{I} \leftarrow i$
    }
    }
\caption{Our Sub-band Selection}
\end{algorithm}

It is worth mentioning that the threshold for selecting the informative sub-bands is chosen using a data-driven method. After sorting the KL-divergence values from highest to lowest, different subsets of sub-bands are selected, e.g., top-5, top-10, and so forth. Suppose that the top-5 values from the set of informative sub-bands are selected. A DNN having input channel size of five is trained on the three training datasets combined, coined the {\it universal dataset}. The performance of the corresponding DNN is reported through Area Under the Curve (AUC) metric using the validation portion of our universal dataset. Fig. 3 depicts the Area Under the Curve (AUC) when different numbers of sub-bands are chosen, based on which the optimal point for number of sub-bands is chosen as 22. The performance of the VISAPP17 dataset is consistent, irrelevant of the number of sub-bands used. This is primarily due to the small size of the VISAPP17 dataset (only 314 images--183 morphed and 131 real), and our DNN easily fits to VISAPP17 dataset regardless of the number of the selected sub-bands.

\section{Experimental Setup}
\subsection{Datasets}
Datasets used in this work are the VISAPP17 \cite{makrushin2017automatic}, MorGAN \cite{damer2018morgan}, and LMA \cite{damer2018morgan}. The VISAPP17 dataset has been created using a landmark-based morphing attack, following by splicing, in which corresponding landmarks in two bona fide subjects are detected and the mean of each pair of the landmarks is calculated. Landmarks of each subject are then warped into the averaged landmark position, and the morphed image is generated using the blending of the two subjects' samples using triangulation \cite{lee1980two} and then spliced into one of the contributing images. This technique aims to avoid artifacts that commonly arise from landmark manipulation, such as those that occur around the hairline. The MorGAN dataset is generated using a GAN. The encoder in a GAN can transform images to a latent space, and when two latent spaces related to two different subjects are combined, a morphed subject is synthesized. The LMA dataset is also generated using the landmark manipulation in two subjects' face images.  

\begin{figure}[h]
\begin{center}
\includegraphics[width=.89\linewidth]{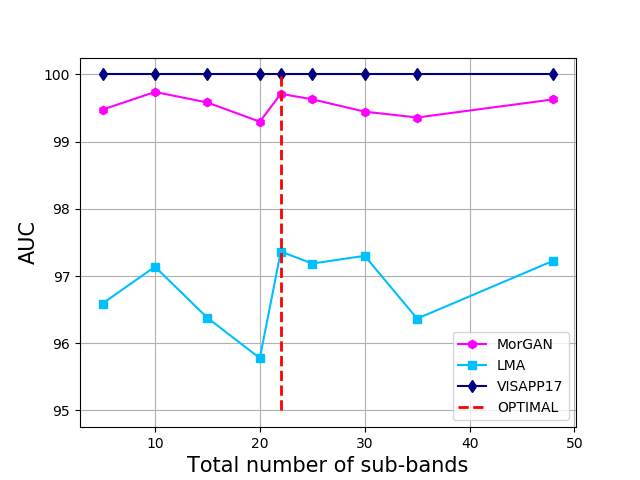} 
\end{center}
  \caption{Area under the curve versus number of sub-bands used in the training. Results indicate that 22 is the optimal number of sub-bands. After selecting 22 sub-bands, the performance does not increase significantly enough to validate using more sub-bands. }
\end{figure}

\subsection{Training Setup}
In this work, the Inception-ResNet-v1 architecture \cite{szegedy2017inception} is adopted as our DNN, which integrates the residual skips introduced in \cite{he2016deep}, and a revised version of Inception architecture \cite{szegedy2016rethinking}. We fine-tune an Inception-ResNet-v1, already pretrained on VGGFace2 \cite{Cao18}. The DNN is additionally fine-tuned with the obtained 22 discriminative wavelet sub-bands of the VISAPP17, MorGAN, and LMA datasets. An Adam optimizer \cite{kingma2014adam} is employed for updating parameters of our network, and two 12 GB TITAN X (Pascal) GPUs accelerate our training.

\begin{table}[t]
\small
\begin{center}
\addtolength{\tabcolsep}{-0pt}
\begin{tabular}{lccccc}
\hline
Train&Test&Algorithm&D-EER&5\%&10\%\\ 
\hline
\multirow{15}{*}{\rotatebox[origin=c]{90}{VISAPP17}}&\multirow{5}{*}{\rotatebox[origin=c]{90}{VISAPP17}}&BSIF+SVM~ \cite{kannala2012bsif}&                      16.51 &35.61&26.79\\
                                &&SIFT+SVM~ \cite{lowe1999object}&     38.59 &82.40&75.60\\
                                &&LBP+SVM~ \cite{ojala1994performance}&  38.00 &77.10 & 67.90\\
                                &&SURF+SVM~ \cite{bay2006surf}&  30.45 &84.70 & 69.40\\
                                &&\textbf{Ours}&                                 {\bf0.00} &{\bf0.00}&{\bf0.00}\\\cline{2-6}
                               
&\multirow{5}{*}{\rotatebox[origin=c]{90}{LMA}}&BSIF+SVM~ \cite{kannala2012bsif}&                      54.00 &93.31&88.95\\
                                &&SIFT+SVM~ \cite{lowe1999object}&     37.00 &79.00&70.00\\
                                &&LBP+SVM~ \cite{ojala1994performance}&  33.00 &71.80 & 59.90\\
                                &&SURF+SVM~ \cite{bay2006surf}&  39.30 &86.10 & 75.70\\
                                &&\textbf{Ours}&                                 {\bf31.86} &{\bf83.80}&{\bf71.21}\\\cline{2-6}

&\multirow{5}{*}{\rotatebox[origin=c]{90}{MorGAN}}&BSIF+SVM~\cite{kannala2012bsif}&                      54.80 &92.32&88.87\\
                                &&SIFT+SVM~ \cite{lowe1999object}&     58.00 &96.10&89.90\\
                                &&LBP+SVM~ \cite{ojala1994performance}&  40.00 &76.90 & 67.40\\
                                &&SURF+SVM~ \cite{bay2006surf}&  40.30 &83.00 & 74.00\\
                                &&\textbf{Ours}&                                 {\bf41.00} &{\bf93.60}&{\bf85.00}\\

\hline
\hline
\multirow{15}{*}{\rotatebox[origin=c]{90}{LMA}}&\multirow{5}{*}{\rotatebox[origin=c]{90}{VISAPP17}}&BSIF+SVM~ \cite{kannala2012bsif}&                      51.19 &83.65&75.00\\
                                &&SIFT+SVM~ \cite{lowe1999object}&     38.00 &90.80&86.30\\
                                &&LBP+SVM~ \cite{ojala1994performance}&  36.60 &77.80 & 71.80\\
                                &&SURF+SVM~ \cite{bay2006surf}&  30.80 &70.00 & 65.60\\
                                &&\textbf{Ours}&                                 {\bf68.80} &{\bf100.00}&{\bf98.90}\\\cline{2-6}
                               
&\multirow{5}{*}{\rotatebox[origin=c]{90}{LMA}}&BSIF+SVM~ \cite{kannala2012bsif}&                      33.05 &78.34&62.86\\
                                &&SIFT+SVM~ \cite{lowe1999object}&     33.30 &83.40&72.00\\
                                &&LBP+SVM~ \cite{ojala1994performance}&  28.00 &58.60 & 51.40\\
                                &&SURF+SVM~ \cite{bay2006surf}&  37.40 &79.50 & 70.00\\
                                &&\textbf{Ours}&                                 {\bf8.80} &{\bf14.90}&{\bf7.91}\\ \cline{2-6}

&\multirow{5}{*}{\rotatebox[origin=c]{90}{MorGAN}}&BSIF+SVM~\cite{kannala2012bsif}&                      42.01 &89.77&79.19\\
                                &&SIFT+SVM~ \cite{lowe1999object}&     50.70 &95.00&89.80\\
                                &&LBP+SVM~ \cite{ojala1994performance}&  35.00 &72.60 & 61.30\\
                                &&SURF+SVM~ \cite{bay2006surf}&  41.27 &84.60 & 78.00\\
                                &&\textbf{Ours}&                                 {\bf32.22} &{\bf76.22}&{\bf62.50}\\

\hline
\hline
\multirow{15}{*}{\rotatebox[origin=c]{90}{MorGAN}}&\multirow{5}{*}{\rotatebox[origin=c]{90}{VISAPP17}}&BSIF+SVM~ \cite{kannala2012bsif}&                      63.00 &100.00&100.00\\
                                &&SIFT+SVM~ \cite{lowe1999object}&     42.00 &92.40&84.00\\
                                &&LBP+SVM~ \cite{ojala1994performance}&  42.32 &84.70 & 79.30\\
                                &&SURF+SVM~ \cite{bay2006surf}&  31.40 &74.00 & 55.70\\
                                &&\textbf{Ours}&                                 {\bf2.20} &{\bf0.59}&{\bf0.00}\\\cline{2-6}

&\multirow{5}{*}{\rotatebox[origin=c]{90}{LMA}}&BSIF+SVM~ \cite{kannala2012bsif}&                      53.00 &95.25&92.46\\
                                &&SIFT+SVM~ \cite{lowe1999object}&     40.20 &90.70&80.00\\
                                &&LBP+SVM~ \cite{ojala1994performance}&  39.18 &75.90 & 67.7\\
                                &&SURF+SVM~ \cite{bay2006surf}&  39.40 &81.00 & 71.60\\
                                &&\textbf{Ours}&                                 {\bf39.11} &{\bf89.55}&{\bf80.25}\\ \cline{2-6}

&\multirow{5}{*}{\rotatebox[origin=c]{90}{MorGAN}}&BSIF+SVM~\cite{kannala2012bsif}&                      1.57 &1.42&1.30\\
                                &&SIFT+SVM~ \cite{lowe1999object}&     43.50 &93.20&84.20\\
                                &&LBP+SVM~ \cite{ojala1994performance}&  20.10 &52.70 & 32.30\\
                                &&SURF+SVM~ \cite{bay2006surf}&  39.95 &80.00 & 72.60\\
                                &&\textbf{Ours}&                                 {\bf0.00} &{\bf0.00}&{\bf0.00}\\

\hline
\hline
\end{tabular}
\end{center}
\caption[Table caption text]{Performance of single morph detection: D-EER\%, BPCER@APCER=5\%, and BPCER@APCER=10\%.} 
\label{table:results_cross}
\end{table}
\subsection{Training/Testing Using Selected Sub-bands }
In order to find the optimal number of sub-bands, we combine the three morph image datasets into a {\it universal dataset}. From this universal dataset, the training set consists of 1631 bona fide, and 1183 morphed samples. The validation set consists of 462 bona fide, and 167 morphed subjects. Moreover, the test set includes 1631 bona fide, and 1183 morphed images. We train several Inception-ResNet-v1 networks using the training portion of the universal dataset for a different number of chosen wavelet sub-bands. We assess the performance of the trained networks using the validation portion of the universal dataset. In other words, we do a search over the number of wavelet sub-bands, which is the input channel size of our convolutional neural network. Please note that the wavelet sub-bands are already sorted based on the corresponding KL-divergence values from highest to lowest. According to the sub-band selection scheme mentioned in section 3.B. and Fig. 3, which shows the performance of the trained classifier using different number of wavelet sub-bands, the optimal number of informative sub-bands is 22; thus, our final DNN has 22 input channels consisting of the top 22 most discriminative sub-bands. 

 The performance of our morph detector, and the baseline methods for comparison are summarized in the Table 1. Please note that we have considered all the possible training/testing scenarios using the three datasets, i.e., the VISAPP17, LMA, and MorGAN. The corresponding Detection Error Trade-off (DET) curves are displayed in Fig. 4. In addition, we have trained the morph classifier using the training portion of the universal dataset, and the performance of that network is also evaluated using the testing portion of each individual dataset, as well as the universal dataset. The results of the training using the universal dataset, and the corresponding baseline methods are provided in the Table 2. Moreover, the related DET curves are shown in Fig. 5. 
 
 D-EER represents Detection Equal Error Rate, where APCER equals BPCER. BPCER5 designates BPCER rate for APCER=5\%, and BPCER10 designates BPCER rate for APCER=10\%. A close scrutiny of the DET curves in Fig. 4 reveals that our morph detector can accurately detect morphed samples in both the VISAPP17, and MorGAN datasets when both training and testing data originate from the same dataset. Fig 5. also shows that our morph detector is able to detect most of the morphed samples in the VISAPP17, and MorGAN datasets when the classifier is trained on the training portion of the universal dataset.

\begin{figure}[t]
\begin{center}
\includegraphics[width=.9\linewidth]{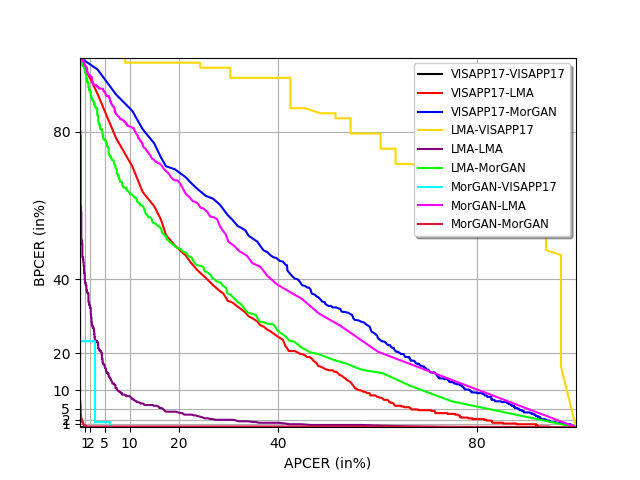} 
\end{center}
   \caption{DET curves when our morph detector is trained and tested on the selected 22-sub-band datasets. The legend represents train-test datasets.}
\end{figure}

\comment{
The DET curve in Fig. 8. shows the performance of our algorithm in different known and unknowns scenarios. In addition, test phase is achieved on the universal dataset while using 22-sub-band data (see Fig. 9). Furthermore, performance of our algorithm is compared to baseline methods benchmarked in table 2. The DNN learns the Known-MORGAN and Known-VISAPP very well with an EER rate of 0.0\% and Known LMA EER at 8\%. Notably, the DNN performs well on unseen data. Unknown MorGAN achieves an Equal Error Rate of 32\%. These results are on par with the state-of-the-art. Table 2 also shows that the LMA morph dataset which was the most difficult database for our model to learn. Fig. 8 demonstrates how sub-band selection has positively boosted morphed imagery detection. The red curve in Fig. 8, that is to say known LMA, shows the performance of the trained classifier when train and test sets are both created using the LMA morphing attack scheme, which are available in the MorGAN dataset. The black curve, being unknown MOR, depicts the performance of the classifier when the test set consists of GAN-based morphed images in the MorGAN dataset while the training set comes from the LMA portion of the MorGAN dataset.}

\begin{table}[t]
\small
\begin{center}
\addtolength{\tabcolsep}{-0pt}
\begin{tabular}{lccccc}
\hline
Train&Test&Algorithm&D-EER&5\%&10\%\\ 
\hline
\multirow{20}{*}{\rotatebox[origin=c]{90}{Universal(VISAPP17+LMA+MorGAN)}}&\multirow{5}{*}{\rotatebox[origin=c]{90}{VISAPP17}}&BSIF+SVM~ \cite{kannala2012bsif}&                      35.00 &67.20&59.00\\
                                &&SIFT+SVM~ \cite{lowe1999object}&     27.00 &83.20&70.90\\
                                &&LBP+SVM~ \cite{ojala1994performance}&  37.67 &72.50 & 59.50\\
                                &&SURF+SVM~ \cite{bay2006surf}&  31.00 &79.40 & 70.10\\
                                &&\textbf{Ours}&                                 {\bf0.00} &{\bf0.00}&{\bf0.00}\\\cline{2-6}
                               
&\multirow{5}{*}{\rotatebox[origin=c]{90}{LMA}}&BSIF+SVM~ \cite{kannala2012bsif}&                      30.00 &70.42&57.60\\
                                &&SIFT+SVM~ \cite{lowe1999object}&     28.31 &67.70&50.00\\
                                &&LBP+SVM~ \cite{ojala1994performance}&  29.00 &61.50 & 51.20\\
                                &&SURF+SVM~ \cite{bay2006surf}&  33.40 &74.50 & 62.70\\
                                &&\textbf{Ours}&                                 {\bf8.61} &{\bf12.93}&{\bf7.05}\\\cline{2-6}

&\multirow{5}{*}{\rotatebox[origin=c]{90}{MorGAN}}&BSIF+SVM~\cite{kannala2012bsif}&                      28.80 &62.42&45.70\\
                                &&SIFT+SVM~ \cite{lowe1999object}&     47.60 &92.30&88.60\\
                                &&LBP+SVM~ \cite{ojala1994performance}&  31.20 &62.00 & 55.60\\
                                &&SURF+SVM~ \cite{bay2006surf}&  38.67 &76.00 & 70.00\\
                                &&\textbf{Ours}&                                 {\bf3.10} &{\bf2.04}&{\bf3.89}\\\cline{2-6}

&\multirow{5}{*}{\rotatebox[origin=c]{90}{Universal}}&BSIF+SVM~ \cite{kannala2012bsif}&                      23.74 &51.42&38.67\\
                                &&SIFT+SVM~ \cite{lowe1999object}&     37.21 &87.45&76.71\\
                                &&LBP+SVM~ \cite{ojala1994performance}&  38.80 &91.36 & 83.40\\
                                &&SURF+SVM~ \cite{bay2006surf}&  36.00 &75.50 & 65.76\\
                                &&\textbf{Ours}&                                 {\bf5.45} &{\bf5.70}&{\bf3.19}\\\cline{2-6}

\end{tabular}
\end{center}
\caption[Table caption text]{Performance of single morph detection: D-EER\%, BPCER@APCER=5\%, and BPCER@APCER=10\%.}  
\label{table:results_cross}
\end{table}


\subsection{Class Activation Maps}
Class activation maps, set forth in \cite{zhou2016learning}, show the extent to which different regions in a given image contribute to the final classification decision for every class in an already trained DNN. After training an Inception-ResNet-v1 morph detector, class activation maps were constructed using the feature embeddings from the last layer before fully connected and softmax layers. The results are interestingly indicative of the likelihood that an image will be classified as morphed or bona fide. For example, in Fig. 6, the middle image, representing a morphed one, has many more affected areas than the other two bona fide images. This is an indicator that the middle image is the most likely image among the three images to be classified as morphed. Given that, our trained DNN using 22 discriminative wavelet sub-bands is effectively distinguishing morphed images from the non-morphed images. 

\begin{figure}[t]
\begin{center}
\includegraphics[width=.9\linewidth]{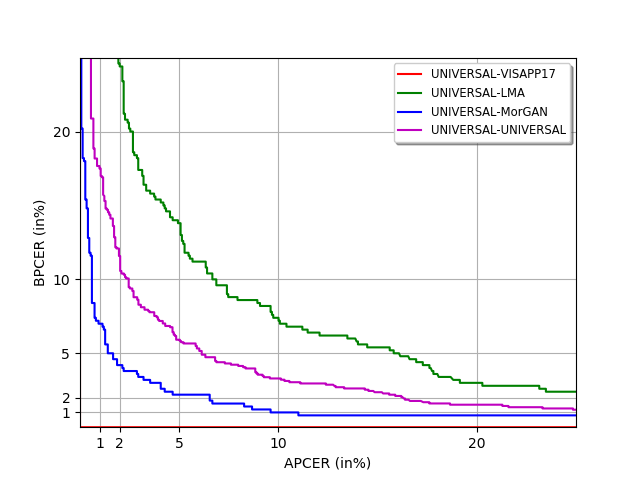} 
\end{center}
   \caption{DET curves when our morph detector is trained on the selected 22-sub-band universal datasets.}
\end{figure}

\begin{figure}[t]
\begin{center}
\includegraphics[width=.85\linewidth]{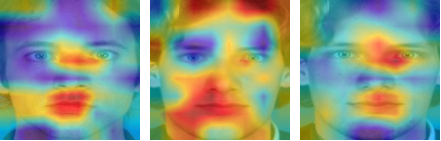} 
\end{center}
   \caption{Class activation maps. Left: bona fide subject 1, middle: morphed subject, right: bona fide subject 2.}
\end{figure}

\section{Ablation Study}
In this section, the effect of sub-band selection is examined. To prove the effectiveness of band selection, a visualization method, namely t-SNE \cite{maaten2008visualizing}, is adopted. A total of 200 morphed, and 200 bona fide images are selected from the test set of MorGAN dataset. Fig. 7 shows the t-SNE visualizations for three scenarios using the MorGAN dataset, the first of which visualizes the original images, which is shown in the leftmost column.  In the middle column, the 48 selected sub-band data is plotted. Finally, the 22 selected sub-band data is shown in the rightmost column. It is evident in Fig. 7 that sub-band selection contributes considerably to concentrating the morphed and bona fide data into separable clusters, which is highly desirable in terms of detecting morphed imagery.





\begin{figure}[t]
\begin{center}
\includegraphics[width=.89\linewidth]{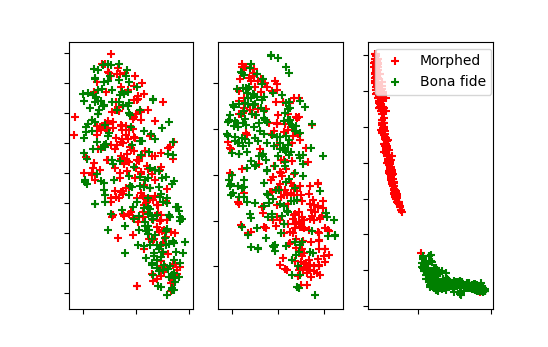} 
\end{center}
   \caption{T-SNE visualization for the original images (left), 48 sub-band data (middle), and 22 sub-band data (right). The 22 sub-bands evidently separate the morph and bona fide classes into very distinct clusters.}
\end{figure}

\section{Conclusion}

In this paper, we proposed a framework to detect morphed face images using undecimated 2D-DWT. To select the optimal and informative bands, we found the distribution of the entropy for all the 48 wavelet sub-bands considering both the bona fide, and morphed images. The KL-divergence between the given distributions, integrated in a data-driven approach, led us to select the 22 most discriminative sub-bands. Furthermore, a close look at the presented results in Tables 1 \& 2 highlights the fact that our morph classifier can identify morphed samples with a high accuracy in both the VISAPP17, and MorGAN datasets. Moreover, the ablation study on the sub-band selection substantiates the effectiveness of our method and shows that our trained DNN can map data samples to a new space where two bona fide and morphed classes are aggregated into two well-separated clusters.

{\small
\bibliographystyle{IEEEtran}
\bibliography{IEEEexample}
}

\end{document}